\documentclass[10pt,twocolumn,letterpaper]{article}

\makeatletter
\def\thanks#1{\protected@xdef\@thanks{\@thanks
        \protect\footnotetext{#1}}}
\makeatother

\usepackage{iccv}
\usepackage{times}
\usepackage{epsfig}
\usepackage{graphicx}
\usepackage{amsmath}
\usepackage{amssymb}

\usepackage{booktabs}
\usepackage{multirow}
\usepackage{diagbox}
\usepackage{makecell}

\usepackage[breaklinks=true,bookmarks=false,colorlinks,citecolor=green]{hyperref}

\iccvfinalcopy 

\def\iccvPaperID{****} 

\ificcvfinal\fi

\begin{document}

\title{Unsupervised Domain Adaptive Detection with Network Stability Analysis}

\author{Wenzhang Zhou$^{1,3*}$, Heng Fan$^{2,*}$, Tiejian Luo$^{1,3}$, Libo Zhang$^{1,3,\dag}$\\
$^1$Institute of Software, Chinese Academy of Sciences, Beijing, China\\
$^2$Department of Computer Science and Engineering, University of North Texas, Denton, USA\\
$^3$University of Chinese Academy of Sciences, Beijing, China\\ 
\thanks{$^{*}$The two authors make equal contributions and are co-first authors.}
\thanks{$^{\dag}$Corresponding author: Libo Zhang (libo@iscas.ac.cn).}
}

\maketitle
\ificcvfinal\thispagestyle{empty}\fi

\begin{abstract}
   Domain adaptive detection aims to improve the generality of a detector, learned from the labeled source domain, on the unlabeled target domain. In this work, drawing inspiration from the concept of \emph{stability} from the control theory that a robust system requires to remain consistent both externally and internally regardless of disturbances, we propose a novel framework that achieves unsupervised domain adaptive detection through stability analysis. In specific, we treat discrepancies between images and regions from different domains as disturbances, and introduce a novel simple but effective \textbf{Network Stability Analysis} (NSA) framework that considers various disturbances for domain adaptation. Particularly, we explore three types of perturbations including heavy and light image-level disturbances and instance-level disturbance. For each type, NSA performs \emph{external consistency analysis} on the outputs from raw and perturbed images and/or \emph{internal consistency analysis} on their features, using teacher-student models. By integrating NSA into Faster R-CNN, we immediately achieve state-of-the-art results. In particular, we set a new record of $52.7\%$ mAP on Cityscapes-to-FoggyCityscapes, showing the potential of NSA for domain adaptive detection. It is worth noticing, our NSA is designed for general purpose, and thus applicable to one-stage detection model (\eg, FCOS) besides the adopted one, as shown by experiments. Code is released at \url{https://github.com/tiankongzhang/NSA}.
\end{abstract}

\section{Introduction}

Benefited by deep neural networks~\cite{krizhevsky2017imagenet,DBLP:conf/corr/SimonyanZ14a,DBLP:conf/cvpr/HeZRS16}, object detection has witnessed considerable progress in recent years \cite{DBLP:conf/eccv/LawD18, DBLP:conf/iccv/Lin2017, DBLP:journals/pami/RenHG017, DBLP:conf/iccv/TianSCH19, DBLP:journals/corr/abs-1904-07850,DBLP:conf/iccv/Girshick_2015}. Modern detectors are usually trained and tested on large-scale annotated datasets \cite{DBLP:journals/ijcv/Mark10, DBLP:conf/eccv/TianSCH19}. Despite excellence, they easily degenerate when applied to images from a new target domain, which heavily limits their practical applications. To mitigate this, a naive solution is to collect a dataset for the new target domain to re-train a detector. Nevertheless, dataset creation is a nontrivial task that needs a large amount of labor. Besides, a new target domain could be arbitrary and it is impossible to collect datasets for all new target domains. To deal with this, researchers explore unsupervised domain adaptation (UDA) detection, aiming to transfer knowledge learned from an annotated source domain to an unlabeled target domain.
\begin{figure}[t]
\centering
\includegraphics[width=0.97\linewidth]{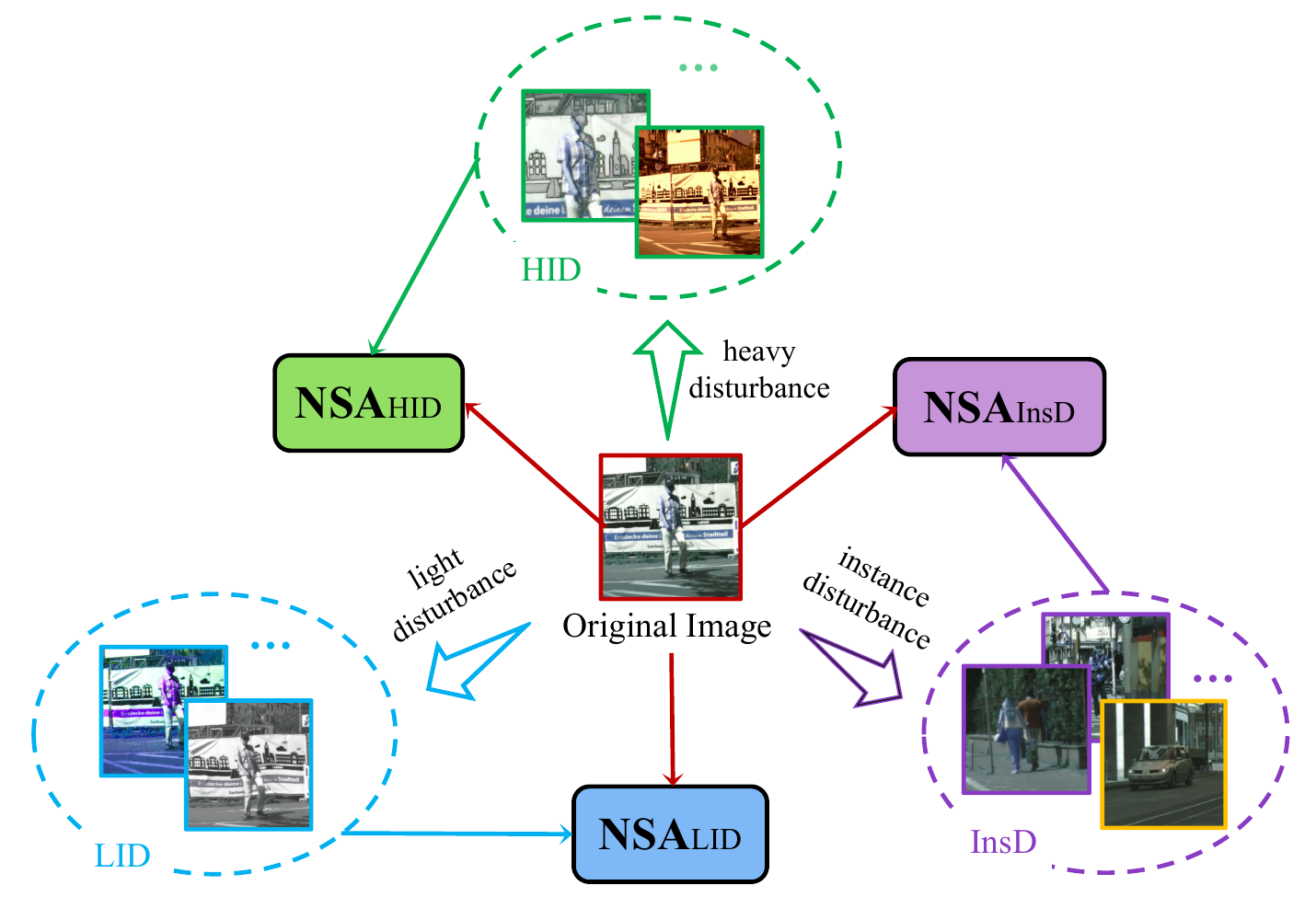}
\caption{Illustration of the proposed NSA framework that applies the specially designed NSA$_{\mathrm{HID}}$, NSA$_{\mathrm{LID}}$ and NSA$_{\mathrm{InsD}}$ for different disturbances including HID, LID and InsD, respectively.}
\label{fig:disturb}
\vspace{-0.9em}
\end{figure}

Existing UDA detection can be generally classified into three families. The first branch focuses on aligning feature distributions of different domains to reduce their gap using, \eg, adversarial learning \cite{DBLP:conf/cvpr/Chen0SDG18, DBLP:conf/cvpr/SaitoUHS19, DBLP:conf/cvpr/ZhuPYSL19} and maximum mean discrepancy \cite{DBLP:conf/icml/pmlr-v37-long15, DBLP:conf/nips/NIPS2016_ac627ab1, DBLP:conf/icml/Mingsheng2017}. Despite effectiveness, these approaches may suffer from three limitations. First, they usually require \emph{both} source and target datasets for training, restraining their usage. Besides, they are problematic with \emph{local misalignment}, principally because of unknown internal distribution of feature space distribution due to the lack of target domain annotations. Finally, a large amount of useful information among samples for different domain datasets is ignored, resulting in inferior performance. Another line leverages self-training for UDA detection~\cite{DBLP:conf/cvpr/RoyChowdhury_2019_CVPR,DBLP:conf/iccv/Khodabandeh_2019_ICCV,DBLP:conf/iccv/Kim_2019_ICCV}. The core idea is to generate high-quality pseudo labels on the target domain and apply them for detector training. Although this strategy improves detection on the target domain, it heavily relies on initial detection results, making them unstable. The third direction is to exploit the teacher-student model~\cite{DBLP:conf/cvpr/Cai_2019_CVPR, DBLP:conf/cvpr/Deng_2021_CVPR}. Using consistency constraints of detector predictions regardless of external disturbance on input, these approaches exhibit robust domain adaptive detection. Despite this, they ignore the consistency for internal features and external predictions under different disturbances, resulting in degradation on the new target domain.

\vspace{0.7em}
\noindent
\textbf{Contributions.} Different than the above methods, we study UDA detection from a new  perspective. Particularly, we observe that the changes of attributes (\eg, scale, view, translation, color) for object and styles for instances are the major causes for domain differences. For a desired stable detector, both feature representations and prediction results should be consistent under these changes. Drawing inspiration from \emph{stability} concept in control theory~\cite{DBLP:book/Bacciotti05} where \emph{the good system needs to perform consistently in both external predictions and internal status in presence of disturbances}, we propose a novel framework for UDA detection via Network Stability Analysis (NSA). The key idea is that, we regard discrepancy caused by distribution changes between two domains as data disturbance, and analyze influence of various disturbances on internal features and external predictions.

More specifically, in this paper we consider three types of disturbances, Heavy and Light Image-level Disturbances (HID and LID) and Instance-level Disturbance (InsD), that involve general perturbations of color, view, texture, scale, translation and instance style in the images. The reason for disturbance division is that variations in images are significantly different and it is difficulty to use a single disturbance for analysis. For each type of disturbance, NSA performs \emph{external consistency analysis} (ECA) on outputs of the original and the disturbed images and/or \emph{internal consistency analysis} (ICA) on their features, both with teacher-student models. Considering each disturbance focuses on different aspects, the NSA is different, accordingly. Concretely, HID majorly focuses on large object or region variations in scales and views. Since the internal features greatly vary while the external detection results are coincident, we only perform ECA in NSA$_{\mathrm{HID}}$ (\ie, NSA for HID). Different from HID, LID mainly contains slight scale and view changes in objects and small pixel displacements, and the local semantics in the internal feature maps are highly similar. Thus, we perform both ECA and ICA in NSA$_{\mathrm{LID}}$ (\ie. NSA for LID). InsD describes differences in instances belonging to the same category. Intuitively, objects of the same class may have adjacent spatial distributions. Inspired by this, we perform ICA in NSA$_{\mathrm{InsD}}$ (\ie, NSA for InsD). Specifically, with real and pseudo labels, we build an undirected graph based on pixel or instance features and further acquire the feature centers of all classes, which exist in an image batch, and select negative samples for each node in the undirected graph from the background region. Finally, the stable feature distribution for all classes is learned with a contrastive loss function. Fig. \ref{fig:disturb} illustrates our idea.

By integrating our NSA of different disturbances into the popular Faster R-CNN \cite{DBLP:journals/pami/RenHG017}, we immediately achieve state-of-the-art results on multiple benchmarks (i.e, Cityscapes \cite{DBLP:conf/cvpr/CordtsORREBFRS16}, FoggyCityscapes \cite{DBLP:journals/ijcv/SakaridisDG18}, RainCityscapes \cite{DBLP:/conf/cvpr/Hu_2019_CVPR}, KITTI \cite{DBLP:/conf/cvpr/are12}, Sim10k \cite{DBLP:/conf/icra/driving17} and BDD100k \cite{DBLP:journals/corr/Yu2018BDD100KAD}), revealing the great potential of NSA for domain adaptive detection. Note that, NSA is designed for general purpose. We show this by plugging NSA into the one-stage detector (\eg, FCOS~\cite{DBLP:conf/iccv/TianSCH19}), and results demonstrate promising performance.

In summary, our contributions are as follows: (\textbf{i}) we propose a novel unified Network Stability Analysis (NSA) framework for domain adaptive detection;
(\textbf{ii}) we introduce the external consistency analysis (ECA) and internal consistency analysis (ICA) for NSA; and (\textbf{iii}) we integrate our NSA for different disturbances into existing detectors and consistently achieve state-of-the-art results.

\section{Related Work}

\noindent
\textbf{Object Detection.} Deep object detection has been greatly advanced~\cite{krizhevsky2017imagenet,DBLP:conf/corr/SimonyanZ14a,DBLP:conf/cvpr/HeZRS16} in recent years. Currently, the modern detectors can be generally categorized into two- or one-stage architectures. The two-stage detectors (\eg, R-CNN \cite{DBLP:conf/cvpr/Girshick_2014} and Fast/Faster R-CNN \cite{DBLP:conf/iccv/Girshick_2015, DBLP:journals/pami/RenHG017}) first extract proposals from an image and then perform classification and regression on these proposals to achieve detection. Because of the excellent results, two-stage framework has been extensively studied with many extensions \cite{DBLP:conf/cvpr/Cai_2018, DBLP:conf/cvpr/Lu_2019}. Different from the two-stage framework, one-stage detectors (\eg, YOLO~\cite{redmon2016you}, CornerNet~\cite{DBLP:conf/eccv/LawD18} and FCOS~\cite{DBLP:conf/iccv/TianSCH19}) remove the proposal stage and directly output object category and location. In this work, we apply Faster R-CNN \cite{DBLP:journals/pami/RenHG017} as our base detector for its outstanding performance, but show generality of our NSA for one-stage detection frameworks.

\begin{figure*}[!t]
\centering
\includegraphics[width=0.95\linewidth]{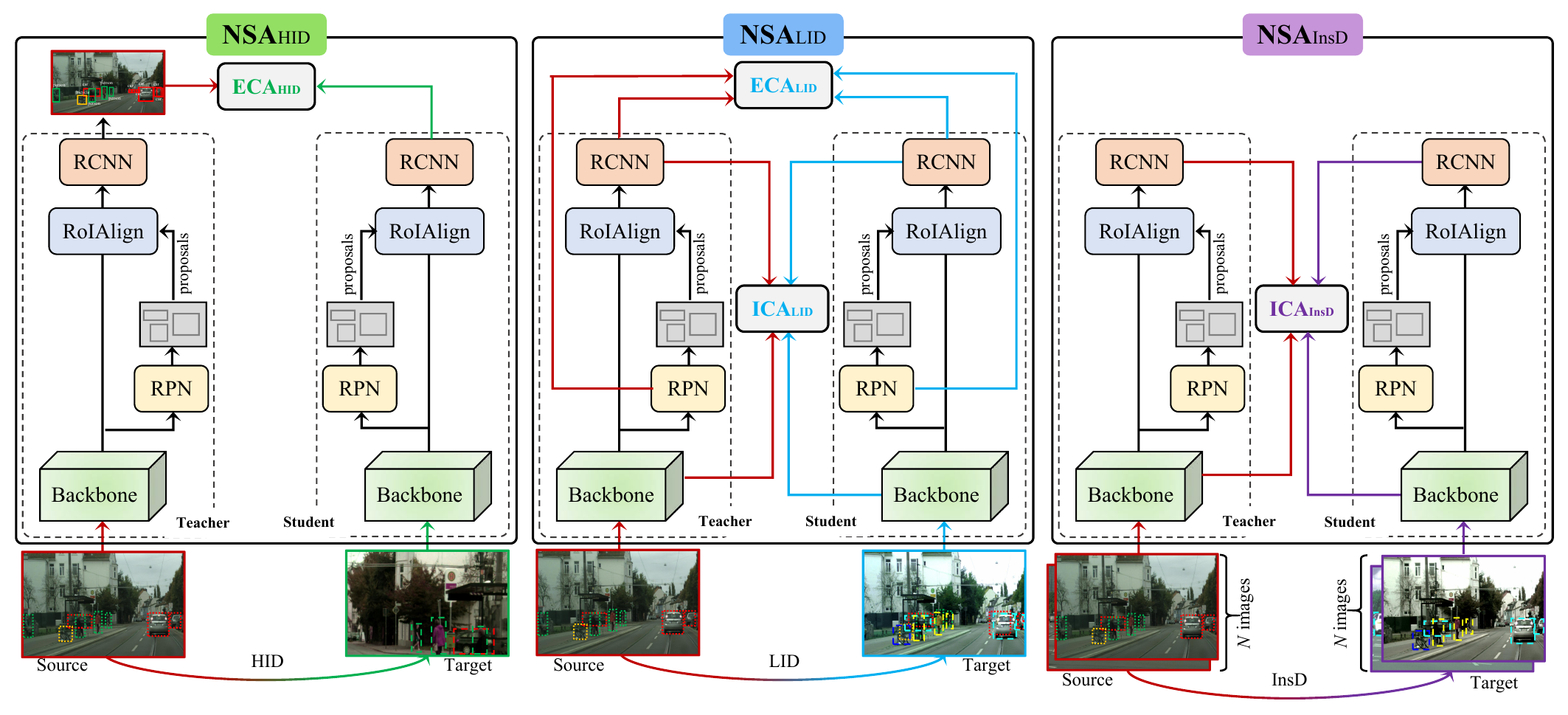}
\caption{Network Stability Analysis (NSA) on different disturbances for UDA detection. Left: We perform NSA$_{\mathrm{HID}}$ to ensure  consistency of detections in images from different domains for HID. Middle: We perform NSA$_{\mathrm{LID}}$ to analyze consistencies of inside features and outside predictions for different images with LID. Right: We perform NSA$_{\mathrm{InsD}}$ by using proximity principle to model feature distribution of instances of the same category or similar regions in InsD. The dashed rectangles in images (bottom) represent objects under disturbances.}
\label{fig:architecture}
\vspace{-1.0em}
\end{figure*}

\vspace{0.2em}
\noindent
\textbf{UDA Detection.} UDA detection aims at improving performance of a detector, trained on the labeled source domain, on the new target domain. Due to its importance, numerous approaches have been proposed. One trend is to align the feature distribution with adversarial learning. The main idea is to design an effective discriminator on various feature spaces, including image-level~\cite{DBLP:conf/eccv/LiDZWLWZ20}, pixel-level~\cite{DBLP:conf/cvpr/KimJKCK19,DBLP:conf/wacv/hsu2020,DBLP:conf/eccv/HsuTLY20}, instance-level~\cite{DBLP:conf/cvpr/Chen0SDG18, DBLP:conf/eccv/SuWZTCQW20, DBLP:journals/tmm/Dayan22} and category-level~\cite{DBLP:conf/cvpr/XuZJW20,DBLP:conf/cvpr/VS_2021,DBLP:conf/cvpr/Zhao_2022}, for detection. Recently, some works \cite{DBLP:conf/cvpr/jiang2021,DBLP:conf/cvpr/Zhou_2022} explore the alignment of fine-grained feature distribution based on combination of multi-levels and effectively reduce the distribution differences between source and target domains.  Despite improvements, they ignore the possible misalignment caused by noise pixels or instances, especially in the background region, or noisy pseudo labels. Besides, another popular line is to adopt self-training to generate pseudo labels on target domain for retraining detector~\cite{DBLP:conf/cvpr/XuWNTZ20,DBLP:conf/cvpr/RoyChowdhury_2019_CVPR,DBLP:conf/iccv/Khodabandeh_2019_ICCV,DBLP:conf/iccv/Kim_2019_ICCV,DBLP:conf/iccv/Zhu_2017_ICCV,DBLP:conf/cvpr/Deng_2021_CVPR}.
However, these methods heavily depend on initial detection results. In this work, we study UAD by analyzing network stability, which significantly differs from above methods.

\vspace{0.2em}
\noindent
\textbf{Consistency Learning for UDA detection.} Consistency-based learning aims to handle the consistent problem under different perturbations. The methods of~\cite{DBLP:conf/cvpr/Kyriazi21,DBLP:conf/nips/NEURIPS2019_d0f4dae8} apply consistency learning on network external predictions. The work of~\cite{DBLP:conf/cvpr/Xie_2021} explores pixel-level consistency for internal feature representation learning. Inspired by this, researchers introduce consistency learning into UDA detection by considering it as a consistency problem of two domains. These approaches are called teacher-student models. The approach of \cite{DBLP:conf/cvpr/Deng_2021_CVPR} leverages the unbiased mean teacher model to reduce the discrepancies in different domains for detection. The method of \cite{DBLP:conf/iccv/Ramamonjison_2021} introduces a simple data augmentation technique named DomainMix with teacher-student model to learn domain-invariant representations and shows excellent results. AT \cite{DBLP:conf/cvpr/li2022cross} uses domain adversarial learning and weak-strong
data augmentation to reduce domain gap. PT \cite{DBLP:conf/icml/chen2022learning} presents a probabilistic teacher to obtain  uncertainty of unlabeled target data with an evolving teacher, and trains the student network in a mutually beneficial manner.

\vspace{0.2em}
\noindent
\textbf{Differences from other works.} In this work, we propose NSA for UDA detection. Our method is related to but different from the above consistency learning or teacher-student methods for UDA detection. \textbf{First}, we consider consistency constraints in both external outputs and internal feature representations while others mainly focus on constraints in one of external model predictions and internal feature. \textbf{Second}, we explore effective network stability analysis method under various and general disturbances while existing methods only study one kind and their performance may degenerate in complex scenarios. In general, our NSA method is a unified solution on what and how to apply consistency on various disturbances for UDA detection.

\section{NSA-based UDA (NSA-UDA) Detection}


\subsection{Overall NSA-UDA Framework}

Fig.~\ref{fig:architecture} shows the overall framework of NSA-UDA. As in Fig.~\ref{fig:architecture}, given an  image $x$, we first apply three disturbances, \ie, HID, LID and InsD (as described later), on $x$ to obtain  perturbed images $\{ x_k \}_{k\in \mathcal{D}}$, where $\mathcal{D}=\{\text{HID},\text{LID}, \text{InsD} \}$. Afterward, we perform NSA for each case. Mathematically, we describe all disturbances with a unified model,
\begin{equation}
    \mathcal{L}_{\text{NSA-UDA}} = \mathcal{L}_{\text{det}}+\sum_{k\in \mathcal{D}}{\gamma_{k} \mathcal{L}_{\text{NSA}_k}(x, x_k)}
\label{equ:gl_model}
\end{equation}
where $\mathcal{L}_{\text{det}}$ denotes the loss of the base student detector as explained later, and $\mathcal{L}_{\text{NSA}_k}$ the loss of $\text{NSA}_k$. $\gamma_{k}$ is a weight to balance the loss. For $\text{NSA}_k$, it contains ECA and/or ICA. Without losing generality, $\mathcal{L}_{\text{NSA}_k}$ can be written as follows,
\begin{equation}
    \mathcal{L}_{\text{NSA}_k}(x, x_k) =  \mathcal{L}_{\text{NSA}_k}^{\text{ECA}}(x, x_k) 
    + \mathcal{L}_{\text{NSA}_k}^{\text{ICA}}(x, x_k) 
\label{equ:gl_dist}
\end{equation}
where $\mathcal{L}_{\text{NSA}_k}^{\text{ECA}}$ and $\mathcal{L}_{\text{NSA}_k}^{\text{ICA}}$ denote the losses for ECA and ICA under disturbance $k \in \mathcal{D}$.

\vspace{0.3em}
\noindent\textbf{Base Detection Architecture.} In this work, teacher or student detector is defined as the base detection. As in Eq. (\ref{equ:gl_model}), $\mathcal{L}_{\text{det}}$ is base student detection loss. In this work, we leverage the two-stage Faster R-CNN~\cite{DBLP:journals/pami/RenHG017} as our base detector for identifying object category and regressing its box. However, it is worth noticing that, the one-stage detector such as FOCS~\cite{DBLP:conf/iccv/TianSCH19} could also be used as the base detector, as shown in our experiments. In general, the detection loss $\mathcal{L}_{\text{det}}$ can be expressed as follows,
\begin{equation}
    \begin{array}{ll}
    \mathcal{L}_{\text{det}}(x,\widehat{y}) = \mathcal{L}_{\text{det}}^{\text{cls}}(x,\widehat{y}) + \mathcal{L}_{\text{det}}^{\text{reg}}(x,\widehat{y})
    \end{array}
\label{equ:det}
\end{equation}
where $\mathcal{L}_{\text{det}}^{\text{cls}}$ and $\mathcal{L}_{\text{det}}^{\text{reg}}$ are the classification and regression loss functions, respectively. $\widehat{y}$ represents the labels of the source domain or pseudo-labels of the target domain.

\subsection{NSA with Disturbance}

\vspace{0.2em}
\noindent
In this work, we regard the discrepancies of domain distributions as input disturbances, and analyze the stability of networks under different disturbances using teacher-student model, aiming at decreasing the impact of disturbances for achieving UDA detection.  In specific, given an image $x$, teacher detector parameterized with $\theta_t$ (\ie, Faster R-CNN) and student detector parameterized with  $\theta_s$ that has identical architecture of teacher detector, we conduct stability analysis NSA$_{\mathrm{HID}}$, NSA$_{\mathrm{LID}}$ and NSA$_{\mathrm{InsD}}$, \emph{externally} and \emph{internally}, for disturbances HID, LID and InsD, respectively.

\subsubsection{NSA$_{\text{HID}}$ for Heavy Image-level Disturbance}

\noindent
\textbf{Heavy Image-level Disturbance} (or \textbf{HID}). HID represents \emph{large} object changes in view and scale with random texture and color variations. To obtain these changes in heavy disturbance, we employ a few common transformation strategies such as random resize, random horizontal flip, center crop, color and texture enhancement to simulate them, where the scale changes randomly in the range $[1,S_{\text{HID}}]$ ($S_{\text{HID}}$ is empirically set to 3.5) and two states of the view change are provided, \ie, $V_{\text{HID}}=1$ and $V_{\text{HID}}=0$, which indicate the image with and without random horizontal flip, respectively. An example of the image with HID can be seen in Fig.~\ref{fig:architecture} (bottom left). Please refer to more examples and pseudo code of HID in supplementary material.

\vspace{0.3em}
\noindent
\textbf{NSA$_{\text{HID}}$}. NSA$_{\text{HID}}$ aims to ensure \emph{externally} consistent and stable predictions for the detector under heavy image-level disturbances in object scales and views. We formulate the ECA of NSA$_{\text{HID}}$ as follows,
\begin{equation}
    \begin{array}{ll}
    \mathcal{L}_{\text{NSA}_{\text{HID}}}^{\text{ECA}}(x, x_{\text{HID}}) = \mathcal{L}_{\text{det}}(x_{\text{HID}}, \widehat{y}, \theta_s)
    \end{array}
\label{equ:eca_hid}
\end{equation}
where $\theta_s$ denotes the parameters of the student detector, and $\widehat{y}$ is the source domain labels or target domain pseudo-labels obtained by the teacher detector.

Since for HID, it is \emph{difficult} to \emph{internally} analyze the consistency on feature maps due to large displacement of pixel-level features, we do not perform the ICA in NSA$_{\text{HID}}$. Thus, we can obtain $\mathcal{L}_{\text{NSA}_{\text{HID}}}^{\text{ICA}}(x, x_{\text{HID}})=0$.

By plugging $\mathcal{L}_{\text{NSA}_{\text{HID}}}^{\text{ECA}}$ and $\mathcal{L}_{\text{NSA}_{\text{HID}}}^{\text{ICA}}$ into Eq. (\ref{equ:gl_dist}), we can compute $\mathcal{L}_{\text{NSA}_\text{HID}}(x, x_\text{HID})$. Fig.~\ref{fig:architecture} (left) illustrates NSA$_\text{HID}$.

\subsubsection{NSA$_{\text{LID}}$ for Light Image-level Disturbance}

\noindent
\textbf{Light Image-level Disturbance} (or \textbf{LID}). LID represents object variations in \emph{small} scale and translation with random texture and color variations, which are simulated by some data transformation strategies in the experiment. Specifically, the scale changes randomly in $[1, S_{\text{LID}}]$ ($S_{\text{LID}}$ is empirically set to 1.5). For translation, we utilize deviation degree, defined by ratio of offset distance and stride of feature block, for measurement and randomly set its value from $[0, D_{\text{LID}}]$ ($D_{\text{LID}}$ is empirically set to 0.25). An example of image with LID is shown in Fig.~\ref{fig:architecture} (bottom middle), and please refer to more examples in supplementary material.

\vspace{0.3em}
\noindent
\textbf{NSA$_{\text{LID}}$}. NSA$_{\text{LID}}$ aims to explore both \emph{external} and \emph{internal} consistency regulations with ECA and ICA, respectively.

The ECA is used for consistency analysis on prediction results, and mathematically formulated as follows,

\vspace{-0.9em}
\begin{equation}
    \footnotesize
    \begin{split}
    \! \mathcal{L}_{\text{NSA}_{\text{LID}}}^{\text{ECA}}(x,x_{\text{LID}}) \! = \! \! \! \! & \sum_{l,k}^{L_{\text{ep}}, C{\text{ep}}} \frac{||A_{l}^{\text{pix}}(O_{l,k}^{\text{pix}}(x,\theta_t)\!-\!O_{l,k}^{\text{pix}}(x_{\text{LID}},\theta_s))||_{2}}{||A_{l}^{\text{pix}}||_{1}} \\
    & \! \! \! \! \! \! \! \! + \varrho \sum\limits_{l=0,k}\limits^{L_{\text{ei}},C_{\text{ei}}}{\frac{||A_{l}^{\text{ins}}(O_{l,k}^{\text{ins}}(x,\theta_t)\!-\!O_{l,k}^{\text{ins}}(x_{\text{LID}},\theta_s))||_{2}}{||A_{l}^{\text{ins}}||_{1}}}
    \end{split}
\label{equ:lid_eca}
\end{equation}

where $O_{l}^{\text{pix}}(\cdot)$ and $O_{l}^{\text{ins}}(\cdot)$ are prediction results generated from teacher or student detectors at pixel and instance levels, respectively. $L_{\text{ep}}$ and $L_{\text{ei}}$ are the numbers of external prediction layers $O^{\text{pix}}$ and $O^{\text{ins}}$, respectively. $C_{\text{ep}}$ and $C_{\text{ei}}$ respectively indicate the set of external prediction categories at pixel- and instance-levels,  \ie, \{`\emph{class}', `\emph{box}'\} in Faster-RCNN or \{`\emph{class}', `\emph{box}' and `\emph{centerness}'\} in FCOS. The indicator $\varrho$ is binary: 1 for the adoption of an instance-level prediction head in the detector (\ie, Faster R-CNN); 0 otherwise (\ie, FCOS). $A_{l}^{\text{pix}}$ represents the weight coefficient of each pixel in prediction maps from the $l^{\text{th}}$ layer, and $A_{l}^{\text{ins}}$ is the weight vector of instances. For the foreground pixels and instances, their weights are 1, otherwise 0. Thus, $A_{l}^{\text{pix}}$ and $A_{l}^{\text{ins}}$ are obtained as follow,
\begin{equation} 
A_{l}^{\text{pix}} = \begin{cases}
    1.0, M_{l}^{\text{pix}} > 0\\
    0.0, \text{otherwise}\\
    \end{cases} \;
A_{l}^{\text{ins}} = \begin{cases}
    1.0, M_{l}^{\text{ins}} > 0\\
    0.0, \text{otherwise}\\
    \end{cases}
    \label{equ:weights_ld}
\end{equation}
where $M_{l}^{\text{pix}}$ and $M_{l}^{\text{ins}}$ are respectively class matrix for pixels and vector for instances from labels and pseudo-labels. For each pixel in $M_{l}^{pix}$ (or each instance in $M_{l}^{ins}$), it is assigned with the class label if belonging to foreground object based on the label (\ie, `$>0$'), otherwise 0.

\begin{figure}[!t]
\centering
\includegraphics[width=\linewidth]{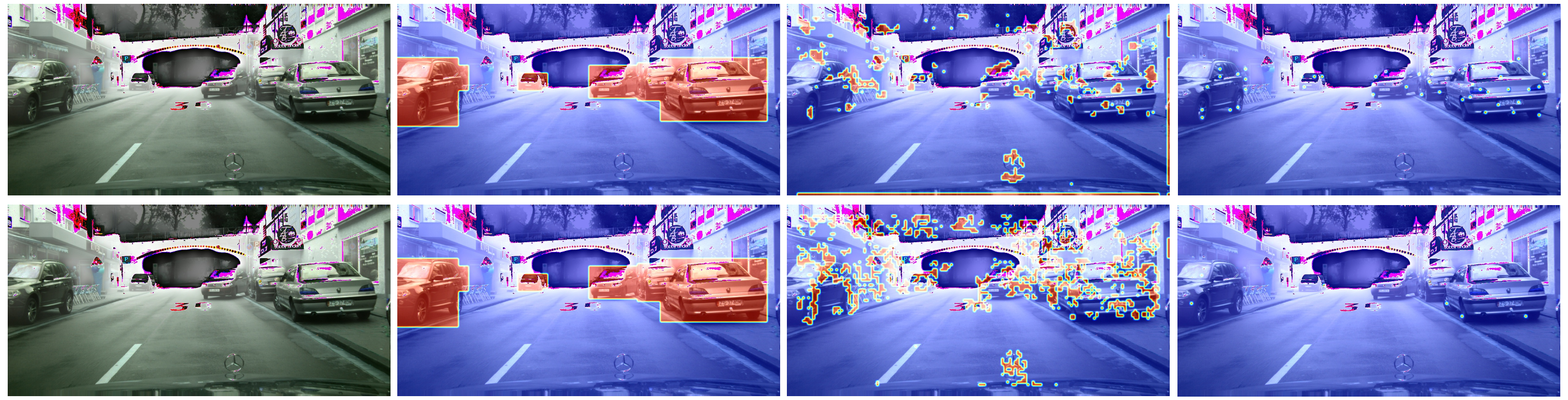}
\caption{Visualization of $A_l^{p}$, $W_l^{t}$ and $B_l^{p}$ on unlabeled target domain using Faster R-CNN detector under Cityscapes-to-FoggyCityscapes adaptation. The first and second rows show the attention areas of weights for $W_{t}=1.0$ and $W_{t}=0.1$ using feature maps after the $3^{\text{th}}$ (\ie, $l=3$) block in backbone. From left to right, they are the original image and heat maps of $A_3^{p}$, $W_3^{t}$ and $B_3^{p}$. We can observe that $A_3^{p}$ mainly focuses on the foreground objects, $W_3^{t}$ on the local textures and $B_3^{p}$ on the sampling  points of objects, as expected.}
\label{fig:weight_vis}
\vspace{-1.0em}
\end{figure}

Different from ECA, ICA is applied for the consistency analysis on feature maps, and expressed as follows,
\begin{equation}
    \footnotesize
    \begin{split}
    \! \! \! \mathcal{L}_{\text{NSA}_{\text{LID}}}^{\text{ICA}}(x,x_{\text{LID}})  = & \sum\limits_{l=0}\limits^{L_{\text{ip}}}{\frac{||B_{l}^{\text{pix}}(F_{l}^{\text{pix}}(x,\theta_t)-F_{l}^{\text{pix}}(x_{\text{LID}},\theta_s))||_{2}}{||B_{l}^{\text{pix}}||_{1}}} + \\
    &  \varrho \sum\limits_{l=0}\limits^{L_{\text{ii}}}{\frac{||B_{l}^{\text{ins}}(F_{l}^{\text{ins}}(x,\theta_t)-F_{l}^{\text{ins}}(x_{\text{LID}},\theta_s))||_{2}}{||B_{l}^{\text{ins}}||_{1}}}
    \end{split}
\label{equ:lid_ica}
\end{equation}
where $L_{\text{ip}}$ and $L_{\text{ii}}$ denote the numbers of pixel-level internal feature layers $F_l^{\text{pix}}$ and instance-level internal feature layers $F_l^{\text{ins}}$. $F_{l}^{\text{pix}}(\cdot)$ and $F_{l}^{\text{ins}}(\cdot)$ are feature maps and vectors generated from teacher or student detectors. $B_{l}^{\text{pix}}$ is the weight coefficient of each pixel in feature maps, and $B_{l}^{\text{ins}}$ denotes the weight vector of instances.

For $B_{l}^{\text{pix}}$, we aim to increase the weights of edges or local contour areas, especially for foreground objects, and meanwhile reduce the interference of abundant smooth patches. To such end, we first estimate the smoothness of local texture as follows,
\begin{equation}
    \begin{array}{ll}
     S_{i,j} = ||F_{i,j}(\theta_t) - H_{i,j}(F(\theta_t), r)||_{1}
    \end{array}
\label{equ:eta3}
\end{equation}
\begin{equation}
    \begin{array}{ll}
     S = \mathcal{R}(S)
    \end{array}
\label{equ:eta4}
\end{equation}
where $H_{ij}(F(\theta_t), r)$ represents the average value of a $r \times r$ window centered at $(i,j)$ on feature $F$ obtained from teacher detector. $\mathcal{R}(\cdot)$ denotes the normalization operation using maximum and minimum values. Next, the local texture is divided into the three categories according to the smoothness, and we assign the different weights to the three types of local texture. After this, we divide the local texture into three kinds according to the smoothness, and assign different weights to each type as follows,
\begin{equation} 
W_t = \begin{cases}
    1.0, \;\;\; S \in (\eta_{2}\overline{s}, \infty]\\
    0.1, \;\;\; S \in (\eta_{1}\overline{s}, \eta_{2}\overline{s}]\\
    0.0, \;\;\; S \in [0, \eta_{1}\overline{s}]
    \end{cases}
\label{tab_wight_wt}
\end{equation}
Here $\eta_{1}$ and $\eta_{2}$ are constant coefficients, and $\overline{s}$ is the average value of $S$. Finally, $B_l^{\text{pix}}$ can be obtained by merging $W_{t}$ and $A_l^{\text{pix}}$ and sampling center points of local areas using by $\varPsi$, 
\begin{equation} 
B_{l}^{\text{pix}} = \varPsi(W_{t} \cdot A_{l}^{\text{pix}}, S_{l})
\end{equation}
Here $\varPsi(\cdot, \cdot)$ represents the operation where center points of local areas are sampled using consistent constraints between the value of center point and the maximum value in a sliding window with stride $1$ on $S_{l}$ map. As displayed in Fig.~\ref{fig:weight_vis}, we visualize $A_l^{\text{pix}}$, $W_t$ and $B_l^{\text{pix}}$ on unlabeled target domain using Faster R-CNN detector under Cityscapes-to-FoggyCityscapes adaptation. 

In $B_{l}^{\text{ins}}$, for the foreground instances, the weights are set to 1, otherwise 0. Thus, $B_{l}^{\text{ins}}$ is obtained by a formula similar to the Eq.\ref{equ:weights_ld}.

By plugging Eq. (\ref{equ:lid_eca}) and (\ref{equ:lid_ica}) into Eq. (\ref{equ:gl_dist}), we can compute $\mathcal{L}_{\text{NSA}_\text{LID}}(x, x_\text{LID})$. Fig.~\ref{fig:architecture} (middle) illustrates NSA$_\text{LID}$.

\subsubsection{NSA$_{\text{InsD}}$ for Instance-level Disturbance}

\noindent
\textbf{Instance-level Disturbance} (or \textbf{InsD}). InsD is an important disturbance in the detection task. It represents variations of objects of the same class in style, scale and view. 

\vspace{0.3em}
\noindent
\textbf{Instance Graph}. To learn a stable UDA detector, in InsD we explore the relation among different instances on feature maps. Specifically, we first extract the instance-level features of objects and background region features to build an instance graph $\mathcal{G}(V,E,D)$ on each of feature layers, where $V \in \mathbb{R}^{N_{g}}$, $E\in \mathbb{R}^{N_{g}\times N_{g}}$, and $D\in \mathbb{R}^{N_{g}\times (C+N_{b})}$ represent the nodes, edges and distances from those nodes to the center of each category and each of $N_b$ samples of background areas in the feature space. $N_{g}$ is the number of nodes, and $C+N_{b}$ includes $C$ classes of the foreground objects and $N_{b}$ background samples that are similar to foreground objects. For the pixel-level feature maps, we use the sliding window strategy and 
the conditions of areas of objects within a certain range and $W_t = 1$ to obtain instance-level features of objects and background region features as nodes.

Assume that $F_{i}$ and $F_{j}$ denote instance-level feature vectors of nodes $V_i$ and $V_j$ in $\mathcal{G}$, the edge $E_{i,j}$ is computed as 
\begin{equation}
    E_{i,j} = 1 - \langle \frac{F_{i}(\theta_s)}{||F_{i}(\theta_s)||_{2}}, \frac{F_{j}(\theta_t)}{||F_{j}(\theta_t)||_{2}} \rangle
\label{equ:eta51}
\end{equation}
where $\langle \cdot, \cdot \rangle$ denotes the dot product function. Then, the $N_{b}$ background samples can be obtained by sorting the values of edges. Subsequently, we further acquire the feature centers of $C$ classes by the following formula,
\begin{equation}
    F_{k, ct}(\theta_t) = \frac{\sum_{i}^{N_{g}}{I(k=c_{i})\cdot F_{i}}(\theta_t)}{\sum_{i}^{N_{g}}{I(k=c_{i})}}
\label{equ:eta6}
\end{equation}
where $F_{k, ct}$ is the feature center of $k^{\text{th}}$ class, and $c_{i}$ indicates the class number of $i^{\text{th}}$ node. Based on the above feature centers of $C$ classes and $N_{b}$ background samples, the distance set of $D$ is easy to obtain as follows,
\begin{equation}
    D_{i, k}^{ct} = \langle \frac{F_{i}(\theta_s)}{||F_{i}(\theta_s)||_{2}}, \frac{F_{k,ct}(\theta_t)}{||F_{k,ct}(\theta_t)||_{2}} \rangle
\label{equ:eta7}
\end{equation}
\begin{equation}
    D_{i, j}^{bg} = \langle \frac{F_{i}(\theta_s)}{||F_{i}(\theta_s)||_{2}}, \frac{F_{j,bg}(\theta_t)}{||F_{j,bg}(\theta_t)||_{2}} \rangle
\label{equ:eta8}
\end{equation}
where $D_{i, k}^{ct}$ and $D_{i, k}^{bg}$ represent the distances from $i^{\text{th}}$ node in $\mathcal{G}$ to feature center of $k^{\text{th}}$ class and the node of $j^{\text{th}}$ of $N_{b}$ background samples.

With the instance graph $\mathcal{G}$  illustrated in supplementary material due to limited space, we perform stability analysis for InsD as follows.

\vspace{0.3em}
\noindent
\textbf{NSA$_{\text{InsD}}$}. NSA$_{\text{InsD}}$ focuses on the \emph{internal} consistency on different instances under InsD. The ICA of NSA$_{\text{InsD}}$ is modeled using the contrastive loss as follows,
\begin{equation}\small
\mathcal{L}_{\text{NSA}_{\text{InsD}}}^{\text{ICA}}(x,x_{\text{InsD}}) = -\sum_{m=0}^{L_{\text{ins}}} \frac{\sum_{i=0}^{N_{g}}{W_{\text{InsD}}^m(i)\cdot\text{log}(p_{i}^{m})}}{\sum_{i=0}^{N_{g}}{W_{\text{InsD}}^{m}(i)}} 
\label{insd_ica}
\end{equation}

\begin{equation}\small
    p_{i}^{m} = \frac{\sum_{k=0}^{C}{I(c_{i}^{m}=k) \cdot\text{exp}({D_{i,k}^{ct,m}})}}{\sum_{k=0}^{C}{\text{exp}({D_{i,k}^{ct,m}})} + \sum_{j=0}^{N_{b}}{\text{exp}({D_{i,j}^{bg,m}})}}
\label{equ:eta9}
\end{equation}
where $I(c_{i}^{m}=k)=1$ if $c_{i}^{m}=k$, otherwise 0. $W_{\text{InsD}}^{m}$ is the weights of nodes in $\mathcal{G}^{m}$, and $W_{\text{InsD}}^{m}(i)=1$ if the $i^\text{th}$ node belongs to the foreground object, otherwise 0. $L_{\text{ins}}$ denotes the number of internal feature layers. 

In InsD, since the prediction results of object categories and bounding boxes are pre-determined, the ECA is not necessary. Therefore, we can obtain $\mathcal{L}_{\text{NSA}_{\text{InsD}}}^{\text{ECA}}=0$.

By plugging $\mathcal{L}_{\text{NSA}_{\text{InsD}}}^{\text{ICA}}$ and $\mathcal{L}_{\text{NSA}_{\text{InsD}}}^{\text{ECA}}$ into Eq. (\ref{equ:gl_dist}), we can compute $\mathcal{L}_{\text{NSA}_\text{InsD}}$. Fig.~\ref{fig:architecture} 
 (right) illustrates NSA$_\text{InsD}$.



\subsection{Optimization}
The training process of our NSA-UDA has three stages. In Stage 1 (S1), the teacher network is trained on only the source domain with Eq.~(\ref{equ:det}) with common data augmentations as in~\cite{DBLP:conf/cvpr/Deng_2021_CVPR,DBLP:conf/cvpr/li2022cross}. Then, in Stage 2 (S2), we further train the student network by Eq.~(\ref{equ:gl_dist}) and update the teacher network by exponential moving average (EMA) on only source domain after initializing $\theta_{s}$ with the trained $\theta_{t}$ as follows,
\begin{equation}
   \theta_{t} = \delta\cdot\theta_{t} + (1-\delta)\cdot\theta_{s}
\label{equ:ema} 
\end{equation}
where $\theta_{t}$ and $\theta_{s}$ represent the parameters of the teacher and student networks. $\delta$ is the EMA rate. In the final Stage (S3), the student and teacher networks are optimized by Eq.~(\ref{equ:gl_dist}) and (\ref{equ:ema}) on source and target domain datasets. 



\section{Experiments}

\noindent
{\bf Implementation.} Our proposed NSA-UDA is implemented in PyTorch~\cite{paszke2019pytorch}. We use Faster R-CNN~\cite{DBLP:journals/pami/RenHG017} with VGG16 \cite{DBLP:conf/corr/SimonyanZ14a} pre-trained on ImageNet \cite{DBLP:conf/cvpr/Deng_2009} as the teacher detector to develop our NSA-UDA. Note that, our method is general and we show this by integrating it into another popular one-stage detection framework FCOS~\cite{DBLP:conf/iccv/TianSCH19} with promising results. The optimizer for training our network employs the SGD approach with a momentum of 0.9 and weight decay of 1e-4. The learning rate is set to 3e-4. The $\eta_{1}$ and $\eta_{2}$ in Eq.~(\ref{tab_wight_wt}) are respectively $1.3$ and $1.6$, and $\gamma_{\text{HID}}$, $\gamma_{\text{LID}}$ and $\gamma_{\text{InsD}}$ in Eq.~(\ref{equ:gl_model}) are empirically set to $1.0$, $0.006$ and $0.001$. The EMA rate $\delta$ in Eq.~(\ref{equ:ema}) is 0.97.

\subsection{Experimental Settings and Datasets}


We conduct extensive experiments under four settings.

\noindent
\textbf{Weather adaptation.} For weather adaptation, we use three datasets with various weathers including Cityscapes~\cite{DBLP:conf/cvpr/CordtsORREBFRS16} (\textbf{C}), FoggyCityscapes~\cite{DBLP:journals/ijcv/SakaridisDG18} (\textbf{F}), and RainCityscapes~\cite{DBLP:/conf/cvpr/Hu_2019_CVPR} (\textbf{R}). Cityscapes is a popular scene understanding benchmark with 2,975 images for training and 500 images for validation. FoggyCityscapes and RainCityscapes are synthesized with fog and rain based on Cityscapes. Among them, FoggyCityscapes has the same number of images in training and validation sets as Cityscapes, but RainCityscapes has 9,432 and 1,188 images for training and validation, respectively. In weather adaptation, we perform two groups of experiments by using Cityscapes as the source domain and FoggyCityscapes or RainCityscapes as the target domain, \ie, \textbf{C$\rightarrow$F} and \textbf{C$\rightarrow$R}.

\vspace{0.2em}
\noindent
\textbf{Small-to-Large adaptation.} For small-to-large adaptation, we use Cityscapes~\cite{DBLP:conf/cvpr/CordtsORREBFRS16} as source domain and BDD100k~\cite{DBLP:journals/corr/Yu2018BDD100KAD} (\textbf{B}) as target domain, \ie, \textbf{C$\rightarrow$B}. In specific, we use a subset of BDD100k, which consists of 36,728 training and 5,258 validation images from 8 classes, for the experiment.  

\vspace{0.2em}
\noindent
\textbf{Cross-Camera adaptation.} For the cross-camera adaptation, we leverage KITTI \cite{DBLP:/conf/cvpr/are12} (\textbf{K}), Cityscapes and FoggyCityscapes for our experiments. Similar to Cityscapes, KITTI is a traffic scene dataset containing 7,481 training images. In the experiment, we utilize KITTI as the source domain and Cityscapes or FoggyCityscapes as the target domain, \ie, \textbf{K$\rightarrow$C} and \textbf{K$\rightarrow$F}, and only consider the category of \emph{car} for evaluation as in \cite{DBLP:conf/eccv/LiDZWLWZ20,DBLP:conf/cvpr/Zhou_2022}. 

\renewcommand{\arraystretch}{0.9}
\begin{table}[!t]\small
    \centering
    \caption{Experiments from \textbf{C$\rightarrow$F} using average precision (AP, in \%). Note that, the best two results are  highlighted in \textbf{\textcolor{red}{red}} and \textbf{\textcolor{blue}{blue}} fonts, respectively, for all state-of-the-art comparison tables.}
    \begin{tabular}{rcc}
    \toprule[1.1pt]
    Method & Backbone  & mAP  \\
    \hline 
    Baseline & VGG-16 & 18.8 \\
    \hline
    GPA \cite{DBLP:conf/cvpr/XuWNTZ20} $_\text{[CVPR'2020]}$ & ResNet-50  &39.5\\
    CFFA \cite{DBLP:conf/cvpr/Zheng0LW20} $_\text{[CVPR'2020]}$ & VGG-16  & 38.6 \\
    DSS \cite{DBLP:conf/cvpr/Wang_2021} $_\text{[CVPR'2020]}$ & ResNet-50 & 40.9\\
    D-adapt \cite{DBLP:conf/cvpr/jiang2021} $_\text{[ICLR'2022]}$  & VGG-16   &41.3 \\
    UMT \cite{DBLP:conf/cvpr/Deng_2021_CVPR} $_\text{[CVPR'2021]}$ & VGG-16  &41.7 \\
    MeGA-CDA \cite{DBLP:conf/cvpr/VS_2021} $_\text{[CVPR'2021]}$ &VGG-16  &41.8 \\
    TIA \cite{DBLP:conf/cvpr/Zhao_2022} $_\text{[CVPR'2022]}$ & VGG-16  & 42.3 \\
    SDA \cite{DBLP:journals/corr/Qianyu21} $_\text{[arXiv'2021]}$ & VGG-16  & \textcolor{blue}{45.2} \\
    TDD \cite{DBLP:conf/cvpr/He_2022} $_\text{[CVPR'2022]}$ & VGG-16  & 43.1 \\
    SIGMA \cite{DBLP:conf/cvpr/Li_2022} $_\text{[CVPR'2022]}$ & VGG-16  & 43.5 \\
    \hline
    Baseline w. Data Aug. (Ours) & VGG-16  & 34.2 \\
    NSA-UDA (Ours) & VGG-16 & $\textcolor{red}{52.7}$\\
    \hline
    Oracle (S1) & VGG-16 & 46.7\\
    Oracle (S2) & VGG-16 & 53.0\\
    \toprule[1.1pt]
    \end{tabular}
\label{tab_city_others}
\vspace{-0.8em}
\end{table}

\renewcommand{\arraystretch}{0.9}
\begin{table}[!t]\small
    \centering
    \caption{Experiments from \textbf{C$\rightarrow$R} using AP (\%).}
    \begin{tabular}{rcc}
    \toprule[1.1pt]
    Method & Backbone & mAP  \\
    \hline 
    DA-Faster~\cite{DBLP:conf/cvpr/Chen0SDG18} $_\text{[CVPR'2018]}$ & VGG-16 & 32.8\\
    SCL \cite{DBLP:journals/corr/abs-1911-02559} $_\text{[arXiv'2019]}$ & VGG-16 & 37.3\\
    SDA \cite{DBLP:journals/corr/Qianyu21} $_\text{[arXiv'2021]}$ & VGG-16 &  \textcolor{blue}{41.5} \\
    \hline
    Baseline w. Data Aug. (Ours) & VGG-16& 48.5\\
    NSA-UDA (Ours) & VGG-16 & $\textcolor{red}{58.7}$ \\
    \hline
    Oracle (S1) & VGG-16 & 41.4\\
    Oracle (S2) & VGG-16 & 44.4 \\
    \toprule[1.1pt]
    \end{tabular}
\label{tab_city_rcity}
\vspace{-0.8em}
\end{table}

\vspace{0.1em}
\noindent
\textbf{Synthetic-to-Real adaptation.} For synthetic-to-real adaptation, we use SIM10k \cite{DBLP:/conf/icra/driving17} (\textbf{M}), Cityscapes and FoggyCityscapes for experiments. SIM10k contains 10k images and 8,550 images are used for training and the rest for validation. In this setting, SIM10k is the source domain and Cityscapes or FoggyCityscapes is the target domain, \ie, \textbf{M$\rightarrow$C} and \textbf{M$\rightarrow$F}. Similar to \cite{DBLP:conf/eccv/LiDZWLWZ20}, we conduct the evaluation on the \emph{car} class.

\subsection{State-of-the-art Comparison}

In this section, we report the experimental evaluation results and comparisons. Note, for fair comparisons, all compared methods adopt~\cite{DBLP:journals/pami/RenHG017} as baseline for implementation.

\noindent
\textbf{Evaluation on Weather adaptation.} Tab. \ref{tab_city_others} exhibits the results from \textbf{C$\rightarrow$F}. As shown in Tab.~\ref{tab_city_others}, NSA-UDA achieves the best mAP of $52.7\%$ and outperforms the second best SDA with $45.2\%$ mAP by $7.5\%$. Compared with UMT that leverages teacher-student learning for domain adaptive detection with $41.7\%$ mAP, our method shows clear improvement with $11.0\%$ gains even using a weaker backbone. In addition, compared to our baseline with $34.2\%$ mAP, we obtain $18.5\%$ mAP gains, evidencing the effectiveness of NSA. Tab. \ref{tab_city_rcity} lists the results from \textbf{C $\rightarrow$ R}. As shown, our NSA-UDA obtains the best result with $58.7\%$ mAP, outperforming the second best SDA with $41.5\%$ mAP by $17.2\%$.

\noindent
\textbf{Evaluation on Small-to-Large adaptation.} We display the results from \textbf{C$\rightarrow$B} in Tab.~\ref{tab_city_scale}. As shown in Tab.~\ref{tab_city_scale}, our NSA-UDA obtains the best mAP of $35.5\%$, outperforming the second best PT with $34.9\%$ mAP. Compared with our baseline of $28.5\%$ mAP, we achieve a gain of $7.0\%$, showing the effectiveness of our NSA model.

\noindent
\textbf{Evaluation on Cross-Camera adaptation.} Tab. \ref{tab:tab_kitti_city} exhibits the results and comparison from \textbf{K$\rightarrow$C}. As shown in Tab. \ref{tab:tab_kitti_city}, the proposed NSA-UDA achieves the second performance with $55.6\%$ AP$_{\mathrm{car}}$. PT performs the best with $60.2\%$ mAP score. However, it is worth noting that PT requires pseudo labels on target domain for self-training, while our NSA can improve generality with only labeled source domain. Compared with our baseline of $46.6\%$, we show a gain of $9.0\%$, verifying the effectiveness of our method.

\noindent
\textbf{Evaluation on Synthetic-to-Real adaptation.} In Tab.~\ref{tab:tab_kitti_city}, we report the results from \textbf{M$\rightarrow$C}. Our NSA-UDA obtains the best result with $56.3\%$ AP$_{\mathrm{car}}$. Compare with the second best PT \cite{DBLP:conf/icml/chen2022learning} with $55.1\%$ AP$_{\mathrm{car}}$, we show $1.2\%$ performance gains. Besides, our method significantly improves the baseline from $44.2\%$ to $56.3\%$, showing its advantages.

\renewcommand{\arraystretch}{0.9}
\begin{table}[!t]\small
    \centering
    \caption{Experiments from \textbf{C$\rightarrow$B} using AP (\%).}
    \begin{tabular}{rcc}
    \toprule[1.1pt]
     Method & Backbone &  mAP  \\
    \hline 
    Baseline & VGG-16 & 23.4\\
    \hline
    DA-Faster \cite{DBLP:conf/cvpr/Chen0SDG18} $_\text{[CVPR'2018]}$ & VGG-16 &24.0\\
    SW-Faster \cite{DBLP:journals/corr/Qianyu21} $_\text{[arXiv'2021]}$ & VGG-16 & 25.3 \\
    SW-Faster-ICR-CCR \cite{DBLP:journals/corr/Qianyu21} $_\text{[arXiv'2021]}$ & VGG-16 &  26.9 \\
    TDD \cite{DBLP:conf/cvpr/He_2022} $_\text{[CVPR'2022]}$ & VGG-16 & 33.6 \\
    PT \cite{DBLP:conf/icml/chen2022learning} $_\text{[ICML'2022]}$ & VGG-16 & \textcolor{blue}{34.9} \\
    \hline
    Baseline w. Data Aug. (Ours) & VGG-16 & 28.5 \\
    NSA-UDA (Ours) & VGG-16 & $\textcolor{red}{35.5}$ \\
    \hline
    Oracle (S1) & VGG-16 & 48.2\\
    Oracle (S2) & VGG-16 & 49.1 \\
    \toprule[1.1pt]
    \end{tabular}
\label{tab_city_scale}
\vspace{-0.8em}
\end{table}

\renewcommand{\arraystretch}{0.9}
\begin{table}[!t]\small
  \centering
  \caption{Experiments from \textbf{K/M$\rightarrow$C} using AP$_{\mathrm{car}}$ (\%).}  
  \setlength{\tabcolsep}{5.0pt}
    \begin{tabular}{rcc}
    \toprule[1.1pt]
    Method & Backbone & AP$_{\mathrm{car}}$  \\
    \hline 
    Baseline & VGG-16 &30.2/30.1  \\
    \hline
    DA-Faster \cite{DBLP:conf/cvpr/Chen0SDG18} $_\text{[CVPR'2018]}$ &VGG-16  & 38.5/39.0 \\
    MAF \cite{DBLP:conf/iccv/HeZ19} $_\text{[ICCV'2019]}$ & VGG-16  & 41.0/41.1 \\
    ATF \cite{DBLP:conf/eccv/HeZ20} $_\text{[ECCV'2020]}$ & VGG-16 &42.1/42.8   \\
    SC-DA \cite{DBLP:conf/cvpr/ZhuPYSL19} $_\text{[CVPR'2019]}$ & VGG-16 &42.5/43.0    \\
    SAPNet \cite{DBLP:conf/eccv/LiDZWLWZ20} $_\text{[ECCV'2020]}$ & VGG-16 &43.4/44.9   \\
    TIA \cite{DBLP:conf/cvpr/Zhao_2022} $_\text{[CVPR'2022]}$ & VGG-16 & 44.0/\: -- \: \\
    DSS \cite{DBLP:conf/cvpr/Wang_2021} $_\text{[CVPR'2021]}$ & ResNet-50 &42.7/44.5  \\
    SSD \cite{DBLP:conf/iccv/Rezaeianaran_2021} $_\text{[ICCV'2021]}$ & ResNet-50 &47.6/49.3  \\
    SIGMA \cite{DBLP:conf/cvpr/Li_2022} $_\text{[CVPR'2022]}$ & VGG-16 & 45.8/53.7  \\
    TDD \cite{DBLP:conf/cvpr/He_2022} $_\text{[CVPR'2022]}$ & VGG-16 & 47.4/53.4  \\
    PT \cite{DBLP:conf/icml/chen2022learning} $_\text{[ICML'2022]}$ & VGG-16 & $\textcolor{red}{60.2}$/\textcolor{blue}{55.1}  \\
    \hline
    Baseline w. Data Aug. (Ours) & VGG-16 & 46.6/44.2    \\
    NSA-UDA (Ours) & VGG-16 & \textcolor{blue}{55.6}/\textcolor{red}{56.3}    \\
    \hline
    Oracle (S1) & VGG-16 & 64.9   \\
    Oracle (S2) & VGG-16 & 67.7   \\
    \toprule[1.1pt]
    \end{tabular}
  \label{tab:tab_kitti_city}
  \vspace{-2.0em}
\end{table}%

Please refer to supplementary material for more results.

\subsection{Ablation Study}
\label{abs}

\noindent
\textbf{NSA-UDA with Different Disturbances.} To analyze different disturbances, we experiment our NSA-UDA on \textbf{C$\rightarrow$F} with different disturbances in S2, as shown in Tab. \ref{tab:aba_disturb}. From Tab. \ref{tab:aba_disturb}, we observe that NSA-UDA with HID significantly improves the baseline from $34.2\%$ mAP to $44.2\%$ mAP. When designing NSA-UDA with all three disturbances, we achieve the best performance with $49.6\%$ mAP. In addition,  when applying our three disturbances to another sota method PT \cite{DBLP:conf/icml/chen2022learning} without our NSA strategy, the result of PT is improved to $44.9\%$ mAP compared to the original perturbation with $42.7\%$ mAP, which however is still much lower than our result with $52.7\%$ mAP in S3, fairly evidencing the effectiveness of our NSA.

\vspace{0.2em}
\noindent
\textbf{NSA-UDA with Different Detectors.} We show the generality of NSA by applying it to FCOS~\cite{DBLP:conf/eccv/TianSCH19} and Deformable DETR ~\cite{zhu2021deformable}. As shown in  Tab.~\ref{absdetector}, our NSA-UDA respectively achieves $44.2\%$ mAP on FCOS and $40.9\%$ mAP on Deformable DETR, significantly outperforming the baselines and other methods~\cite{DBLP:conf/eccv/HsuTLY20,DBLP:conf/aaai/Li_2022,DBLP:conf/cvpr/Zhou_2022} on FCOS.

\vspace{0.2em}
\noindent
\textbf{NSA-UDA with Different Training Stages.} To verify the effect of different training stages, we conduct extensive experiments on seven adaptions as displayed in Tab.~\ref{tab:tab_multi_stages}.  From Tab.~\ref{tab:tab_multi_stages}, compared with S1 (\ie, baseline with data augmentation), S2 (\ie, our NSA) can significantly improve performance using only source domain data in all settings, showing the generality of our analysis. When employing target domain pseudo labels, S3 further boosts the performance.

\renewcommand{\arraystretch}{0.9}
\begin{table}[!t]\small
\centering
\caption{NSA-UDA with different disturbances on \textbf{C$\rightarrow$F}.}
\tabcolsep=0.08cm
\begin{tabular}{c|ccccc}
    \toprule[1.1pt]
    Method & TDA(Ours) & NSA$_{\mathrm{HID}}$ & NSA$_{\mathrm{LID}}$ & NSA$_{\mathrm{InsD}}$ & mAP (\%) \\  
    \hline 
     \multirow{5}{*}{NSA-UDA} & \checkmark &     &       &       & 34.2 \\ 
    & \checkmark & \checkmark    &       &       &  44.2\\
    & \checkmark & \checkmark    & \checkmark     &       & 48.9 \\ 
    & \checkmark & \checkmark    &       & \checkmark     & 45.9 \\ 
    & \checkmark & \checkmark    & \checkmark     & \checkmark     & 49.6 \\
    \hline
    \multirow{2}{*}{PT \cite{DBLP:conf/icml/chen2022learning}} & &     &       &       & 42.7 \\
     & \checkmark &     &       &       & 44.9 \\
    \toprule[1.1pt]
    \end{tabular}%
  \label{tab:aba_disturb}%
  \vspace{-0.8em}
\end{table}

\renewcommand{\arraystretch}{0.9}
\begin{table}[!t]\small
\centering
\caption{NSA-UDA with different detectors on \textbf{C$\rightarrow$F}. The backbones for Faster R-CNN/FCOS and Deformable DETR are VGG-16 and ResNet-50.}
\begin{tabular}{rcc}
\toprule[1.1pt]
    Method & Detector  & mAP (\%) \\
    \hline 
    Baseline w. Data Aug. (Ours)  & Faster R-CNN  & 34.2 \\
    NSA-UDA (Ours)  & Faster R-CNN  & 52.7 \\
    \hline 
    CFA \cite{DBLP:conf/eccv/HsuTLY20} $_\text{[ECCV'2020]}$   & FCOS  & 36.0 \\
    SCAN \cite{DBLP:conf/aaai/Li_2022} $_\text{[AAAI'2022]}$ &FCOS  & 42.1 \\
    MGADA \cite{DBLP:conf/cvpr/Zhou_2022} $_\text{[CVPR'2022]}$ &FCOS  & 43.6 \\
    \hline
    Baseline w. Data Aug. (Ours) & FCOS   & 21.0 \\
    NSA-UDA (Ours)  & FCOS   & 44.2 \\ 
    \hline
    Baseline w. Data Aug. (Ours) & Deformable DETR   & 28.5 \\
    NSA-UDA (Ours)  & Deformable DETR   & 40.9 \\ 
    \toprule[1.1pt]
    \end{tabular}%
  \label{absdetector}
  \vspace{-1.0em}
\end{table}

\renewcommand{\arraystretch}{0.9}
\begin{table}[!t]\small
  \centering
  \caption{Effect of training stages on different adaption settings.}
  \setlength{\tabcolsep}{1.6mm}
  {
    \begin{tabular}{@{}cccccccccc@{}}
    \toprule[1.1pt]
    \rule{0pt}{5pt} \multirow{3}{*}{S1} &
    \multirow{3}{*}{S2} &
    \multirow{3}{*}{S3} &
    \rotatebox{270}{\textbf{C$\rightarrow$F}} & \rotatebox{270}{\textbf{C$\rightarrow$R}} & \rotatebox{270}{\textbf{C$\rightarrow$B}} & \rotatebox{270}{\textbf{K$\rightarrow$C}} & \rotatebox{270}{\textbf{K$\rightarrow$F}} & \rotatebox{270}{\textbf{M$\rightarrow$C}} & \rotatebox{270}{\textbf{M$\rightarrow$F}}\\
    \hline
    \checkmark     &       &       & 34.2  & 48.5  & 28.5  & 46.6  & 22.6  & 44.2 & 26.6\\
    \checkmark     & \checkmark     &       & 49.6  & 55.1  & 33.7  & 52.9  & 41.9  & 52.2 & 40.4\\
    \checkmark     & \checkmark     & \checkmark     & 52.7  & 58.7  & 35.5 & 55.6  & 50.0  & 56.3 & 46.0  \\
    \toprule[1.1pt]
    \end{tabular}%
    }
  \label{tab:tab_multi_stages}%
  \vspace{-0.7em}
\end{table}%

\begin{table}[!t]
\begin{minipage}[c]{0.23\textwidth}
\renewcommand{\arraystretch}{0.9}
\small
\centering
\caption{Weight analysis of $\gamma_{LID}$ in $NSA_{LID}$.} 
\tabcolsep=0.05cm
\begin{tabular}{ccccc}
\toprule[1.1pt]
    $\gamma_{LID}$  & 6e-4 & 6e-3 & 6e-2 & 6e-1 \\
    \hline
    mAP (\%)  & 48.3 & 49.6 & 49.2 & 48.7 \\
    \toprule[1.1pt]
    \end{tabular}%
  \label{tab:weight_gamma_lid}%
\end{minipage}
\hspace{0.1em}
\begin{minipage}[c]{0.23\textwidth}
\small
\renewcommand{\arraystretch}{0.9}
\centering
\caption{Weight analysis of $\gamma_{InsD}$ in $NSA_{InsD}$.} 
\tabcolsep=0.05cm
\begin{tabular}{ccccc}
\toprule[1.1pt]
    $\gamma_{LID}$  & 1e-4 & 1e-3 & 1e-2 & 1e-1 \\
    \hline
    mAP (\%)  & 49.1 & 49.6 & 48.3 & 48.2 \\
    \toprule[1.1pt]
    \end{tabular}%
  \label{tab:weight_gamma_insd}%
\end{minipage}
\vspace{-1.2em}
\end{table}

\begin{figure}[!t]
\centering
\includegraphics[width=\linewidth]{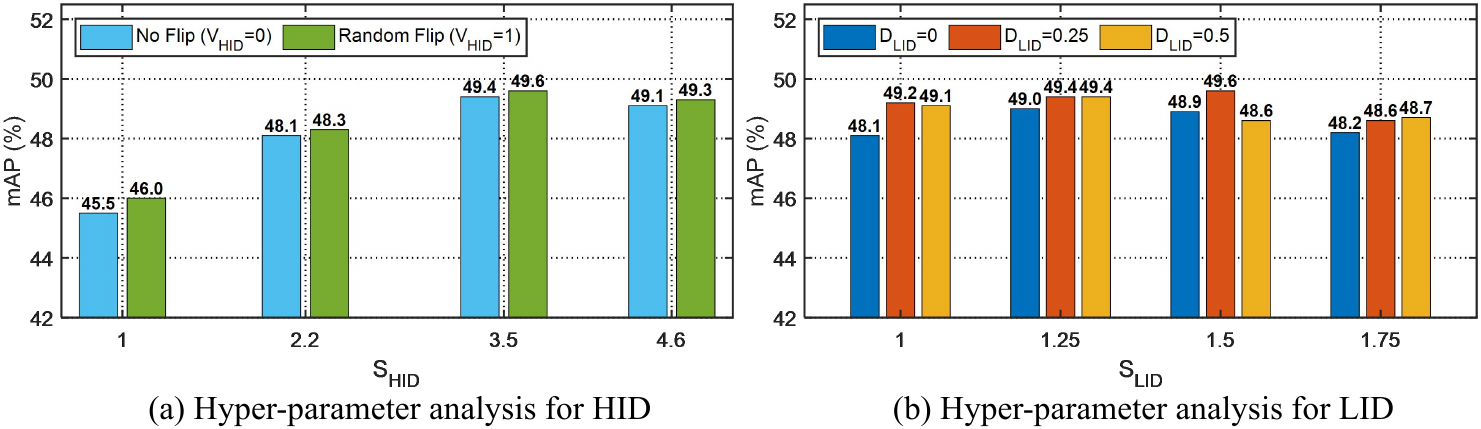}
\caption{Hyper-parameter analysis in disturbance for NSA-UDA.}
\label{fig:pmv_vis}
\end{figure}

\vspace{0.2em}
\noindent
\textbf{NSA-UDA with Different Disturbing Degrees.} To study the impact of disturbances, we conduct comparisons with different hyper-parameters in NSA$_{\text{HID}}$ and NSA$_{\text{LID}}$. As in Fig. \ref{fig:pmv_vis}, HID with relatively large scale variation and flip can improve our performance. In particular, $S_{\text{HID}}=3.5$ with random flip (\ie, $V_{\text{HID}}=1$) achieves the best mAP score of $49.6\%$. For LID, adding small disturbances in scale and displacement can boost the detector, and the best $49.6\%$ mAP score is obtained with  $D_{\text{LID}}=0.25$ and $S_{\text{LID}}=1.5$.

\renewcommand{\arraystretch}{0.9}
\begin{table}[!t]\small
\centering
\caption{Weight analysis of different types of local texture.}
\tabcolsep=0.15cm
\begin{tabular}{ccccc}
\toprule[1.1pt]
    W$_{t}^1$  & 1.0  & 1.0 & 1.0 & 1.0 \\
    W$_{t}^2$  & 0.1  & 1.0 & 0.1 & 1.0 \\
    W$_{t}^3$  & 0.0  & 0.0 & 1.0 & 1.0 \\
    \hline
    mAP (\%)   & 49.6 & 48.6 & 48.8 & 48.1  \\
    \toprule[1.1pt]
    \end{tabular}%
  \label{tab:weight_texture}%
  \vspace{-0.8em}
\end{table}

\renewcommand{\arraystretch}{0.9}
\begin{table}[!t]\small
\begin{minipage}{0.2\textwidth}
\centering
\caption{\small{Analysis of ECA and ICA in NSA$_{\mathrm{LID}}$.}}
\tabcolsep=0.1cm
\begin{tabular}{ccc}
\toprule[1.1pt]
    NSA$_{\mathrm{LID}}^\mathrm{ECA}$ & NSA$_{\mathrm{LID}}^\mathrm{ICA}$  & mAP (\%) \\  
  \hline
          &            & 45.9 \\ 
    \checkmark         &       & 47.2 \\
          & \checkmark         & 48.7 \\ 
    \checkmark        & \checkmark     &  49.6\\ 
    \toprule[1.1pt]
    \end{tabular}%
  \label{tab:ab_eca_ica}%
\end{minipage}
\hspace{5ex} 
\begin{minipage}{0.2\textwidth}
\centering
\caption{Number of local textures in NSA$_{\mathrm{LID}}$.}
\tabcolsep=0.2cm
\begin{tabular}{cc}
\toprule[1.1pt]
    Num. of Types & mAP (\%) \\ 
    \hline 
    1       & 48.0 \\
    2       & 48.5 \\
    3       & 49.6 \\
    4       & 49.0 \\
    \toprule[1.1pt]
    \end{tabular}%
  \label{tab:division_texture}%
\end{minipage}
\vspace{-2.23em}
\end{table}

\vspace{0.2em}
\noindent
\textbf{NSA-UDA with Local-Texture Division.} To study different division and weight assignment for local texture in Eq. (\ref{tab_wight_wt}), we conduct ablations on number of types
for local texture in Tab. \ref{tab:division_texture} and  different weights in Tab. \ref{tab:weight_texture}. 
As shown in Tab. \ref{tab:division_texture}, when number of local textures is three, our NSA achieves the best mAP score of $49.6\%$, demonstrating the necessity and rationality of division of local texture. Meanwhile, Tab. \ref{tab:weight_texture} shows 1/0.1/0.0 achieves satisfying results.

\vspace{0.2em}
\noindent
\textbf{ECA and ICA of LID.} To investigate ECA and ICA in $NSA_{LID}$, we conduct ablations in S2 from \textbf{C$\rightarrow$F} on ECA and ICA in NSA$_{\mathrm{LID}}$ in Tab. \ref{tab:ab_eca_ica}. From Tab .\ref{tab:ab_eca_ica}, We see obvious gains by ECA and ICA, showing their effectiveness. 

\textbf{NSA-UDA with Different Disturbance Weights.} To probe the effect of weights $\gamma$ in Eq.(1) in paper, we conduct ablations in S2 from \textbf{C$\rightarrow$F} for LID in Tab. \ref{tab:weight_gamma_lid} and for InsD in Tab. \ref{tab:weight_gamma_insd}. As shown in Tab .\ref{tab:weight_gamma_lid}, $\gamma_{LID}=0.006$ achieves the best mAP score of $49.6\%$. Larger or smaller value of $\gamma_{LID}$ can reduce the performance of our $NSA_{LID}$. Meanwhile, in  Tab .\ref{tab:weight_gamma_insd}, $\gamma_{InsD}=0.001$ achieves the best mAP of $49.6\%$.


\section{Conclusion}
In this paper, we explore UDA detection from a different perspective. In particular, we regard discrepancies between different domains as disturbances and propose a network stability analysis (NSA) framework for domain adaptive detection under different disturbances. By utilizing NSA on Faster R-CNN, our UDA detector, NSA-UDA, shows state-of-the-art performance on multiple benchmarks. In addition, our NSA is general and  applicable to different detection frameworks.

\vspace{0.3em}
\noindent
{\bf Acknowledgement.} Libo Zhang was supported by Youth Innovation Promotion Association, CAS (2020111). Heng Fan and his employer received no financial support for this work.

{\small
\bibliographystyle{ieee_fullname}
\bibliography{egbib}

\begin{thebibliography}{10}\itemsep=-1pt

\bibitem{DBLP:book/Bacciotti05}
Andrea~Bacciotti andLionel Rosier.
\newblock Liapunov functions and stability in control theory, 2005.

\bibitem{DBLP:conf/cvpr/Cai_2019_CVPR}
Qi Cai, Yingwei Pan, Chong-Wah Ngo, Xinmei Tian, Lingyu Duan, and Ting Yao.
\newblock Exploring object relation in mean teacher for cross-domain detection.
\newblock In {\em CVPR}, 2019.

\bibitem{DBLP:conf/cvpr/Cai_2018}
Zhaowei Cai and Nuno Vasconcelos.
\newblock Cascade r-cnn: Delving into high quality object detection.
\newblock In {\em CVPR}, 2018.

\bibitem{DBLP:conf/icml/chen2022learning}
Meilin Chen, Weijie Chen, Shicai Yang, Jie Song, Xinchao Wang, Lei Zhang,
  Yunfeng Yan, Donglian Qi, Yueting Zhuang, Di Xie, et~al.
\newblock Learning domain adaptive object detection with probabilistic teacher.
\newblock In {\em ICML}, 2022.

\bibitem{DBLP:conf/cvpr/Chen0SDG18}
Yuhua Chen, Wen Li, Christos Sakaridis, Dengxin Dai, and Luc~Van Gool.
\newblock Domain adaptive faster {R-CNN} for object detection in the wild.
\newblock In {\em CVPR}, 2018.

\bibitem{DBLP:conf/cvpr/CordtsORREBFRS16}
Marius Cordts, Mohamed Omran, Sebastian Ramos, Timo Rehfeld, Markus Enzweiler,
  Rodrigo Benenson, Uwe Franke, Stefan Roth, and Bernt Schiele.
\newblock The cityscapes dataset for semantic urban scene understanding.
\newblock In {\em CVPR}, 2016.

\bibitem{DBLP:conf/cvpr/Deng_2021_CVPR}
Jinhong Deng, Wen Li, Yuhua Chen, and Lixin Duan.
\newblock Unbiased mean teacher for cross-domain object detection.
\newblock In {\em CVPR}, 2021.

\bibitem{DBLP:journals/ijcv/Mark10}
Mark Everingham, Luc~Van Gool, Christopher~KI Williams, John Winn, and Andrew
  Zisserman.
\newblock The pascal visual object classes (voc) challenge.
\newblock {\em IJCV}, 88(2):303–338, 2010.

\bibitem{DBLP:/conf/cvpr/are12}
Andreas Geiger, Philip Lenz, and Raquel Urtasun.
\newblock Are we ready for autonomous driving? the {KITTI} vision benchmark
  suite.
\newblock In {\em CVPR}, 2012.

\bibitem{DBLP:conf/iccv/Girshick_2015}
Ross Girshick.
\newblock Fast r-cnn.
\newblock In {\em ICCV}, 2015.

\bibitem{DBLP:conf/cvpr/Girshick_2014}
Ross Girshick, Jeff Donahue, Trevor Darrell, and Jitendra Malik.
\newblock Rich feature hierarchies for accurate object detection and semantic
  segmentation.
\newblock In {\em CVPR}, 2014.

\bibitem{DBLP:journals/tmm/Dayan22}
Dayan Guan, Jiaxing Huang, Aoran Xiao, Shijian Lu, and Yanpeng Cao.
\newblock Uncertainty-aware unsupervised domain adaptation in object detection.
\newblock {\em TMM}, 24:2502--2514, 2021.

\bibitem{DBLP:conf/cvpr/HeZRS16}
Kaiming He, Xiangyu Zhang, Shaoqing Ren, and Jian Sun.
\newblock Deep residual learning for image recognition.
\newblock In {\em CVPR}, 2016.

\bibitem{DBLP:conf/cvpr/He_2022}
Mengzhe He, Yali Wang, Jiaxi Wu, Yiru Wang, Hanqing Li, Bo Li, Weihao Gan, Wei
  Wu, and Yu Qiao.
\newblock Cross domain object detection by target-perceived dual branch
  distillation.
\newblock In {\em CVPR}, 2022.

\bibitem{DBLP:conf/iccv/HeZ19}
Zhenwei He and Lei Zhang.
\newblock Multi-adversarial faster-rcnn for unrestricted object detection.
\newblock In {\em ICCV}, 2019.

\bibitem{DBLP:conf/eccv/HeZ20}
Zhenwei He and Lei Zhang.
\newblock Domain adaptive object detection via asymmetric tri-way faster-rcnn.
\newblock In {\em ECCV}, 2020.

\bibitem{DBLP:conf/eccv/HsuTLY20}
Cheng{-}Chun Hsu, Yi{-}Hsuan Tsai, Yen{-}Yu Lin, and Ming{-}Hsuan Yang.
\newblock Every pixel matters: Center-aware feature alignment for domain
  adaptive object detector.
\newblock In {\em ECCV}, 2020.

\bibitem{DBLP:conf/wacv/hsu2020}
Han-Kai Hsu, Chun-Han Yao, Yi-Hsuan Tsai, Wei-Chih Hung, Hung-Yu Tseng, Maneesh
  Singh, and Ming-Hsuan Yang.
\newblock Progressive domain adaptation for object detection.
\newblock In {\em WACV}, 2020.

\bibitem{DBLP:/conf/cvpr/Hu_2019_CVPR}
Xiaowei Hu, Chi-Wing Fu, Lei Zhu, and Pheng-Ann Heng.
\newblock Depth-attentional features for single-image rain removal.
\newblock In {\em CVPR}, 2019.

\bibitem{DBLP:conf/nips/NEURIPS2019_d0f4dae8}
Jisoo Jeong, Seungeui Lee, Jeesoo Kim, and Nojun Kwak.
\newblock Consistency-based semi-supervised learning for object detection.
\newblock In {\em Advances in Neural Information Processing Systems},
  volume~32, 2019.

\bibitem{DBLP:conf/cvpr/Deng_2009}
Deng Jia, Dong Wei, Socher Richard, Li Li-Jia, Li Kai, and Fei-Fei Li.
\newblock Imagenet: A large-scale hierarchical image database.
\newblock In {\em CVPR}, 2009.

\bibitem{DBLP:conf/cvpr/jiang2021}
Junguang Jiang, Baixu Chen, Jianmin Wang, and Mingsheng Long.
\newblock Decoupled adaptation for cross-domain object detection.
\newblock In {\em ICLR}, 2022.

\bibitem{DBLP:/conf/icra/driving17}
Matthew Johnson{-}Roberson, Charles Barto, Rounak Mehta, Sharath~Nittur
  Sridhar, Karl Rosaen, and Ram Vasudevan.
\newblock Driving in the matrix: Can virtual worlds replace human-generated
  annotations for real world tasks?
\newblock In {\em ICRA}, 2017.

\bibitem{DBLP:conf/iccv/Khodabandeh_2019_ICCV}
Mehran Khodabandeh, Arash Vahdat, Mani Ranjbar, and William~G. Macready.
\newblock A robust learning approach to domain adaptive object detection.
\newblock In {\em ICCV}, 2019.

\bibitem{DBLP:conf/iccv/Kim_2019_ICCV}
Seunghyeon Kim, Jaehoon Choi, Taekyung Kim, and Changick Kim.
\newblock Self-training and adversarial background regularization for
  unsupervised domain adaptive one-stage object detection.
\newblock In {\em ICCV}, 2019.

\bibitem{DBLP:conf/cvpr/KimJKCK19}
Taekyung Kim, Minki Jeong, Seunghyeon Kim, Seokeon Choi, and Changick Kim.
\newblock Diversify and match: {A} domain adaptive representation learning
  paradigm for object detection.
\newblock In {\em CVPR}, 2019.

\bibitem{krizhevsky2017imagenet}
Alex Krizhevsky, Ilya Sutskever, and Geoffrey~E Hinton.
\newblock Imagenet classification with deep convolutional neural networks.
\newblock In {\em NIPS}, 2012.

\bibitem{DBLP:conf/eccv/LawD18}
Hei Law and Jia Deng.
\newblock Cornernet: Detecting objects as paired keypoints.
\newblock In {\em ECCV}, 2018.

\bibitem{DBLP:conf/eccv/LiDZWLWZ20}
Congcong Li, Dawei Du, Libo Zhang, Longyin Wen, Tiejian Luo, Yanjun Wu, and
  Pengfei Zhu.
\newblock Spatial attention pyramid network for unsupervised domain adaptation.
\newblock In {\em ECCV}, 2020.

\bibitem{DBLP:conf/aaai/Li_2022}
Wuyang Li, Xinyu Liu, Xiwen Yao, and Yixuan Yuan.
\newblock Scan: Cross domain object detection with semantic conditioned
  adaptation.
\newblock {\em AAAI}, 36(2), 2022.

\bibitem{DBLP:conf/cvpr/Li_2022}
Wuyang Li, Xinyu Liu, and Yixuan Yuan.
\newblock Sigma: Semantic-complete graph matching for domain adaptive object
  detection.
\newblock In {\em CVPR}, 2022.

\bibitem{DBLP:conf/cvpr/li2022cross}
Yu-Jhe Li, Xiaoliang Dai, Chih-Yao Ma, Yen-Cheng Liu, Kan Chen, Bichen Wu,
  Zijian He, Kris Kitani, and Peter Vajda.
\newblock Cross-domain adaptive teacher for object detection.
\newblock In {\em CVPR}, 2022.

\bibitem{DBLP:conf/iccv/Lin2017}
Tsung-Yi Lin, Priya Goyal, Ross Girshick, Kaiming He, and Piotr Dollar.
\newblock Focal loss for dense object detection.
\newblock In {\em ICCV}, 2017.

\bibitem{DBLP:conf/eccv/TianSCH19}
Tsung-Yi Lin, Michael Maire, Serge Belongie, James Hays, Pietro Perona, Deva
  Ramanan, Piotr Dollar, and C~Lawrence Zitnick.
\newblock Microsoft coco: Common objects in context.
\newblock In {\em ECCV}, 2014.

\bibitem{DBLP:conf/icml/pmlr-v37-long15}
Mingsheng Long, Yue Cao, Jianmin Wang, and Michael Jordan.
\newblock Learning transferable features with deep adaptation networks.
\newblock In {\em ICML}, 2015.

\bibitem{DBLP:conf/nips/NIPS2016_ac627ab1}
Mingsheng Long, Han Zhu, Jianmin Wang, and Michael~I Jordan.
\newblock Unsupervised domain adaptation with residual transfer networks.
\newblock In {\em NIPS}, 2016.

\bibitem{DBLP:conf/icml/Mingsheng2017}
Mingsheng Long, Han Zhu, Jianmin Wang, and Michael~I. Jordan.
\newblock Deep transfer learning with joint adaptation networks.
\newblock In {\em ICML}, 2017.

\bibitem{DBLP:conf/cvpr/Lu_2019}
Xin Lu, Buyu Li, Yuxin Yue, Quanquan Li, and Junjie Yan.
\newblock Grid r-cnn.
\newblock In {\em CVPR}, 2019.

\bibitem{DBLP:conf/cvpr/Kyriazi21}
Luke Melas-Kyriazi and Arjun~K. Manrai.
\newblock Pixmatch: Unsupervised domain adaptation via pixelwise consistency
  training.
\newblock In {\em 2021 IEEE/CVF Conference on Computer Vision and Pattern
  Recognition (CVPR)}, pages 12430--12440, 2021.

\bibitem{paszke2019pytorch}
Adam Paszke, Sam Gross, Francisco Massa, Adam Lerer, James Bradbury, Gregory
  Chanan, Trevor Killeen, Zeming Lin, Natalia Gimelshein, Luca Antiga, et~al.
\newblock Pytorch: An imperative style, high-performance deep learning library.
\newblock {\em NeurIPS}, 2019.

\bibitem{DBLP:conf/iccv/Ramamonjison_2021}
Rindra Ramamonjison, Amin Banitalebi-Dehkordi, Xinyu Kang, Xiaolong Bai, and
  Yong Zhang.
\newblock Simrod: A simple adaptation method for robust object detection.
\newblock In {\em ICCV}, 2021.

\bibitem{redmon2016you}
Joseph Redmon, Santosh Divvala, Ross Girshick, and Ali Farhadi.
\newblock You only look once: Unified, real-time object detection.
\newblock In {\em CVPR}, 2016.

\bibitem{DBLP:journals/pami/RenHG017}
Shaoqing Ren, Kaiming He, Ross~B. Girshick, and Jian Sun.
\newblock Faster {R-CNN:} towards real-time object detection with region
  proposal networks.
\newblock {\em TPAMI}, 39(6):1137--1149, 2017.

\bibitem{DBLP:conf/iccv/Rezaeianaran_2021}
Farzaneh Rezaeianaran, Rakshith Shetty, Rahaf Aljundi, Daniel~Olmeda Reino,
  Shanshan Zhang, and Bernt Schiele.
\newblock Seeking similarities over differences: Similarity-based domain
  alignment for adaptive object detection.
\newblock In {\em ICCV}, 2021.

\bibitem{DBLP:conf/cvpr/RoyChowdhury_2019_CVPR}
Aruni RoyChowdhury, Prithvijit Chakrabarty, Ashish Singh, SouYoung Jin, Huaizu
  Jiang, Liangliang Cao, and Erik Learned-Miller.
\newblock Automatic adaptation of object detectors to new domains using
  self-training.
\newblock In {\em CVPR}, 2019.

\bibitem{DBLP:conf/cvpr/SaitoUHS19}
Kuniaki Saito, Yoshitaka Ushiku, Tatsuya Harada, and Kate Saenko.
\newblock Strong-weak distribution alignment for adaptive object detection.
\newblock In {\em CVPR}, 2019.

\bibitem{DBLP:journals/ijcv/SakaridisDG18}
Christos Sakaridis, Dengxin Dai, and Luc~Van Gool.
\newblock Semantic foggy scene understanding with synthetic data.
\newblock {\em IJCV}, 126(9):973--992, 2018.

\bibitem{DBLP:journals/corr/abs-1911-02559}
Zhiqiang Shen, Harsh Maheshwari, Weichen Yao, and Marios Savvides.
\newblock {SCL:} towards accurate domain adaptive object detection via gradient
  detach based stacked complementary losses.
\newblock {\em arXiv}, 2019.

\bibitem{DBLP:conf/corr/SimonyanZ14a}
Karen Simonyan and Andrew Zisserman.
\newblock Very deep convolutional networks for large-scale image recognition.
\newblock In {\em ICLR}, 2015.

\bibitem{DBLP:conf/eccv/SuWZTCQW20}
Peng Su, Kun Wang, Xingyu Zeng, Shixiang Tang, Dapeng Chen, Di Qiu, and
  Xiaogang Wang.
\newblock Adapting object detectors with conditional domain normalization.
\newblock In {\em ECCV}, 2020.

\bibitem{DBLP:conf/iccv/TianSCH19}
Zhi Tian, Chunhua Shen, Hao Chen, and Tong He.
\newblock {FCOS:} fully convolutional one-stage object detection.
\newblock In {\em ICCV}, 2019.

\bibitem{DBLP:conf/cvpr/VS_2021}
Vibashan VS, Vikram Gupta, Poojan Oza, Vishwanath~A. Sindagi, and Vishal~M.
  Patel.
\newblock Mega-cda: Memory guided attention for category-aware unsupervised
  domain adaptive object detection.
\newblock In {\em CVPR}, 2021.

\bibitem{DBLP:conf/cvpr/Wang_2021}
Yu Wang, Rui Zhang, Shuo Zhang, Miao Li, Yangyang Xia, Xishan Zhang, and Shaoli
  Liu.
\newblock Domain-specific suppression for adaptive object detection.
\newblock In {\em CVPR}, 2021.

\bibitem{DBLP:conf/cvpr/Xie_2021}
Zhenda Xie, Yutong Lin, Zheng Zhang, Yue Cao, Stephen Lin, and Han Hu.
\newblock Propagate yourself: Exploring pixel-level consistency for
  unsupervised visual representation learning.
\newblock In {\em Proceedings of the IEEE/CVF Conference on Computer Vision and
  Pattern Recognition (CVPR)}, pages 16684--16693, June 2021.

\bibitem{DBLP:conf/cvpr/XuZJW20}
Chang{-}Dong Xu, Xing{-}Ran Zhao, Xin Jin, and Xiu{-}Shen Wei.
\newblock Exploring categorical regularization for domain adaptive object
  detection.
\newblock In {\em CVPR}, 2020.

\bibitem{DBLP:conf/cvpr/XuWNTZ20}
Minghao Xu, Hang Wang, Bingbing Ni, Qi Tian, and Wenjun Zhang.
\newblock Cross-domain detection via graph-induced prototype alignment.
\newblock In {\em CVPR}, 2020.

\bibitem{DBLP:journals/corr/Yu2018BDD100KAD}
Fisher Yu, Haofeng Chen, Xin Wang, Wenqi Xian, Yingying Chen, Fangchen Liu,
  Vashisht Madhavan, and Trevor Darrell.
\newblock Bdd100k: A diverse driving dataset for heterogeneous multitask
  learning.
\newblock In {\em CVPR}, 2020.

\bibitem{DBLP:conf/cvpr/Zhao_2022}
Liang Zhao and Limin Wang.
\newblock Task-specific inconsistency alignment for domain adaptive object
  detection.
\newblock In {\em CVPR}, 2022.

\bibitem{DBLP:conf/cvpr/Zheng0LW20}
Yangtao Zheng, Di Huang, Songtao Liu, and Yunhong Wang.
\newblock Cross-domain object detection through coarse-to-fine feature
  adaptation.
\newblock In {\em CVPR}, 2020.

\bibitem{DBLP:journals/corr/Qianyu21}
Qianyu Zhou, Qiqi Gu, Jiangmiao Pang, Zhengyang Feng, Guangliang Cheng, Xuequan
  Lu, Jianping Shi, and Lizhuang Ma.
\newblock Self-adversarial disentangling for specific domain adaptation.
\newblock {\em arXiv}, 2021.

\bibitem{DBLP:conf/cvpr/Zhou_2022}
Wenzhang Zhou, Dawei Du, Libo Zhang, Tiejian Luo, and Yanjun Wu.
\newblock Multi-granularity alignment domain adaptation for object detection.
\newblock In {\em CVPR}, 2022.

\bibitem{DBLP:journals/corr/abs-1904-07850}
Xingyi Zhou, Dequan Wang, and Philipp Kr{\"{a}}henb{\"{u}}hl.
\newblock Objects as points.
\newblock {\em arXiv}, 2019.

\bibitem{DBLP:conf/iccv/Zhu_2017_ICCV}
Jun-Yan Zhu, Taesung Park, Phillip Isola, and Alexei~A. Efros.
\newblock Unpaired image-to-image translation using cycle-consistent
  adversarial networks.
\newblock In {\em ICCV}, 2017.

\bibitem{DBLP:conf/cvpr/ZhuPYSL19}
Xinge Zhu, Jiangmiao Pang, Ceyuan Yang, Jianping Shi, and Dahua Lin.
\newblock Adapting object detectors via selective cross-domain alignment.
\newblock In {\em CVPR}, 2019.

\bibitem{zhu2021deformable}
Xizhou Zhu, Weijie Su, Lewei Lu, Bin Li, Xiaogang Wang, and Jifeng Dai.
\newblock Deformable detr: Deformable transformers for end-to-end object
  detection.
\newblock In {\em ICLR}, 2021.

\end{thebibliography}
}

\end{document}